\documentclass[lettersize,journal]{IEEEtran}

\usepackage{amsfonts}
\usepackage{amsmath} 
\usepackage{multirow}

\usepackage{amsfonts}
\usepackage{booktabs}
\usepackage{siunitx}
\usepackage{graphicx}
\usepackage{gensymb}
\usepackage{tablefootnote}

\usepackage{tabularx}
\usepackage{makecell}
\usepackage{url}
\usepackage[noadjust]{cite}

\usepackage{comment}
\usepackage{soul}
\usepackage{xcolor}
\definecolor{darkgreen}{rgb}{0.0, 0.5, 0.0}

\newcommand{\KM}[1]{{\color{orange}#1}}

\usepackage{amssymb}
\usepackage{pifont}
\newcommand{\xmark}{\ding{55}}

\title{\LARGE \bf{CLRNet}: Targetless Extrinsic Calibration for Camera, Lidar and 4D Radar Using Deep Learning}

\author{Marcell Kegl, Andras Palffy, Csaba Benedek  and Dariu M. Gavrila
\thanks{
This work has been submitted to the IEEE for possible publication. Copyright may be transferred without notice, after which this version may no longer be accessible.
This work was supported in part by the European Union within the framework of the National Laboratory for Autonomous Systems (RRF-2.3.1-21-2022-00002), and also by the TKP2021-NVA-01 and OTKA \#143274 projects of the Hungarian NRDI Office.}
 \thanks{M. Kegl and C. Benedek are with the Hungarian Research Network Institute for Computer Science and Control (HUN-REN SZTAKI), 1111 Budapest, Kende utca 13-17, and with the  P\'azm\'any P\'eter Catholic University, Faculty of Information Technology and Bionics, 1083 Budapest, Pr\'ater utca 50/A. (e-mail: kegl.marcell@sztaki.hun-ren.hu, benedek.csaba@sztaki.hun-ren.hu), D. Gavrila is with the Intelligent Vehicles Group, TU Delft, 2628 CD Delft, Netherlands (e-mail: d.m.gavrila@tudelft.nl). A. Palffy is with Perciv AI, Delft, the Netherlands.
}
}

\begin{document}

\maketitle
\thispagestyle{empty}
\pagestyle{empty}


\begin{abstract}
In this paper, we address extrinsic calibration for camera, lidar, and 4D radar sensors. Accurate extrinsic calibration of radar remains a challenge due to the sparsity of its data. We propose CLRNet, a novel, multi-modal end-to-end deep learning (DL) calibration network capable of addressing joint camera-lidar-radar calibration, or pairwise calibration between any two of these sensors. 
We incorporate equirectangular projection, camera-based depth image prediction, additional radar channels, and leverage lidar with a shared feature space and loop closure loss. 

In extensive experiments using the View-of-Delft and Dual-Radar datasets, we demonstrate superior calibration accuracy compared to existing state-of-the-art methods, reducing both median translational and rotational calibration errors by at least 50\%. Finally, we examine the domain transfer capabilities of the proposed network and baselines, when evaluating across datasets. 
The code will be made publicly available upon acceptance at: \url{https://github.com/tudelft-iv}.
\end{abstract}

\section{Introduction}
\label{seq:introduction}

\IEEEPARstart{A}{ccurate} and robust 3D perception is essential for autonomous systems (e.g. self-driving vehicles, mobile robots) operating in an unknown or dynamic environment. Using optical cameras, we can obtain rich color and texture information, while with lidar and radar sensors, we can measure accurate distance information, independently of lighting conditions. By the fusion of these different sensor modalities, we can leverage their individual advantages to provide an improved, more comprehensive, and robust perception of the environment. However, using data from multiple sensors requires knowledge of their relative pose; therefore, accurate extrinsic calibration is an essential prerequisite for sensor fusion.

To solve this 6 Degrees of Freedom (DoF) problem (3 rotational and 3 translational DoF), some extrinsic calibration methods rely on a dedicated
target object such as a checkerboard \cite{zhang_calib_2004, domhof_calibration}. However, these target-based approaches require time-consuming manual work or a specific calibration environment with preliminary installed marker objects, and they are not suitable for various real-world situations where online or single-shot calibration is needed (e.g., dealing with a non-rigid multi-sensor platform,  or scenes where target objects are unavailable). As a result, automatic targetless calibration becomes increasingly important.


Targetless methods seek to enhance the flexibility and automation of extrinsic calibration by gathering information directly from the environment, eliminating the need for specific target objects \cite{survey_CL}. Traditional targetless approaches are two-stage methods starting with a feature extraction step, after which the parameters are regressed with a traditional numerical optimization method, like gradient descent. 
However, these methods rely on the presence and reliable extraction of feature points \cite{RGKCNet}, edges \cite{hku_calib}, or segmented areas \cite{semalign}.

In contrast, end-to-end Deep Learning (DL) based approaches can potentially be more accurate by jointly optimizing feature extraction and parameter regression. Their performance does, however, depend on the model, the amount of training data, and the domain shift between training and test set (e.g., different sensor characteristics and configurations, various road environments).
Being able to handle domain shifts is important for real-world deployment, where retraining may be impractical \cite{han2025dfcalib, huang2025whatmatters}.  While domain transfer capabilities have recently been explored in some recent camera–lidar calibration studies \cite{han2025dfcalib, huang2025whatmatters, cmrnext, DXQ-Net}, investigations involving radar calibration and generalization across different datasets having been lacking.


In this article, we consider extrinsic calibration in the automotive context involving 4D radar \cite{apalffy2022}. This latest generation radar sensor measures targets in three spatial dimensions, rather than in two, plus Doppler velocity\footnote{In the remainder, we will denote radars involving 3 and 2 spatial dimensions by 4D and 3D, respectively, also accounting for the Doppler dimension}. While several accurate methods exist for camera-to-lidar calibration \cite{LCCNet, CalibDepth, RGGNet}, accurate targetless radar calibration remains an unsolved problem due to the sparsity of radar data \cite{Scholler_radar_DL, 4DRC-OC}.

We introduce a new deep neural network (DNN), called CLRNet, to perform joint targetless calibration in multi-sensor systems containing camera, lidar, and radar, in an end-to-end DL fashion. By considering the three sensors jointly, we can exploit the advantages of established camera-to-lidar calibration techniques, and also the fact that both lidar and radar provide range measurements that can be matched in the depth image domain. As a result, we obtain an improved camera-radar alignment through a shared feature space (containing information from all three sensor modalities) and a joint loop closure term in the loss function. Our experiments show that by handling the three sensors jointly, a more accurate calibration can be achieved between the camera and radar than the available state-of-the-art (SOTA) techniques, and also than performing only pairwise (camera-radar) calibration.

The introduced architecture is flexible, and it can be extended in a straightforward way to jointly calibrate more than three sensors upon availability, or smaller network variants can be applied for pairwise sensor calibration.


Our proposed method is capable of estimating calibration parameters from a single time frame of synchronized measurements from the three sensors. This capability is particularly important for non-rigid sensor platforms, where the relative positions and orientations of the sensors may vary over time due to articulation (e.g., two-wheelers, robotic arms) or vibrations.
In many real-world scenarios, however, the sensor platform can be considered rigid over a short period of time (e.g., sensors on-board intelligent vehicles), allowing calibration to be performed across multiple frames.


Our main contributions are the following:

\begin{itemize}
    \item We propose CLRNet, a network for the joint calibration of camera, lidar, and 4D radar sensors, whose key novelties include the use of a shared feature space
    and the adoption of equirectangular projection for depth map creation.
    CLRNet incorporates previous techniques, such as camera-based depth prediction \cite{CalibDepth}, additional radar channels \cite{apalffy2022}, and loop closure loss \cite{domhof_calibration}, within our multi-modal setup.
    \item We also introduce a network variant called CLRNet+4, which processes multiple time frames as input and outputs an aggregated transformation matrix, for rigid platforms.
    \item  We perform extensive ablation studies to study the effectiveness of such techniques and various architecture design choices, especially with respect to radar calibration.
    Our network outperforms state-of-the-art targetless calibration approaches in radar-related accuracy using the View-of-Delft (VoD) \cite{apalffy2022} and Dual-Radar \cite{Dual-Radar} datasets. Finally, we examine domain transfer capabilities of our network and a baseline across these datasets.
    \item We use open datasets and make our software open-source for non-commercial purposes to increase reproducibility and impact.
\end{itemize}

\section{Related Work}
\label{seq:related_work}

Our work addresses the problem of targetless extrinsic calibration of 4D radar using an end-to-end DNN. In this section, we first provide a brief overview of targetless extrinsic calibration methods in general. We then summarize key developments in the well-studied area of end-to-end deep learning-based calibration between lidar and camera. Finally, we present a more detailed review of targetless calibration approaches that specifically involve 4D radar.

\subsection{Targetless Extrinsic Calibration}

Traditional calibration approaches explicitly reason about geometry and use a cost function to evaluate the quality of the estimated extrinsic parameters, which are obtained by solving an optimization problem.
Survey papers \cite{survey_CL} and \cite{calibration_review} divide these methods into three categories, based on how the necessary geometrical information for the calibration is extracted. There are information-based \cite{pandey_mutual_calib_2012}, feature-based \cite{RGKCNet, hku_calib,semalign} and motion-based \cite{nagyB_2020_calib} approaches.
In addition to the traditional approaches, several \textit{deep learning (DL) based methods} have recently been proposed \cite{LCCNet, CalibDepth, RGGNet,regnet} to perform calibration between lidar and camera. While traditional calibration approaches might also use deep neural networks to extract features before the optimization phase \cite{RGKCNet, semalign}, end-to-end DL methods directly regress the extrinsic parameters.

\subsection{End-to-End Deep Learning for Lidar–Camera Calibration}

An early end-to-end DL based calibration network, called Regnet 
\cite{regnet}, uses Network in Network layers for both feature extraction and matching, followed by fully connected layers to obtain the final parameters. They use iterative refinement, a cascade of multiple networks, which are trained with measurements miscalibrated to varying degrees, an approach later applied by other methods as well \cite{calibration_review}.
CalibNet \cite{CalibNet} increases the calibration accuracy over Regnet by ResNet-based \cite{resnet} feature extraction and using photometric and point cloud distance losses. LCCNet \cite{LCCNet} uses ResNet-18 for feature extraction, switching the feature matching to a PWC-Net \cite{PWC-Net} inspired correlation layer, and optimizing parameter distance loss and point cloud distance loss. As one of the first successful DL based lidar-camera calibration methods (mean average error of \( 1.01 \, \text{cm and } 0.12  \, \text{°} \)), LCCNet inspired several subsequent approaches \cite{CalibDepth, DedgeNet, scnet} that can be considered as its modified or supplemented versions \cite{calibration_review}.

\subsection{Targetless Calibration Involving Radar}

Recent advances in radar technology have led to increasingly dense radar point clouds \cite{Dual-Radar}, yet they remain sparse compared to lidar, posing significant challenges for accurate targetless calibration methods \cite{4DRC-OC}. Moreover, calibrating radars with cameras is challenging since the two sensors lack shared features like color, shape, or depth \cite{Scholler_radar_DL}. These properties, combined with issues of measurement noise, missed detections, and false positives, make it more difficult to accurately calibrate radar sensors. Consequently, there are only a few approaches for automatic and targetless radar-camera calibration \cite{Scholler_radar_DL, 4DRC-OC}. 
An overview of the most relevant existing targetless multi-modal extrinsic calibration approaches involving radar is provided in Table \ref{tab:related_work}, which is elaborated on in the following subsections.

\setlength{\tabcolsep}{3pt} 
\begin{table*}[t]
\centering
\caption{Targetless extrinsic calibration approaches involving radar}
\label{tab:related_work}
\resizebox{\textwidth}{!}{
\begin{tabular}{@{}llcccclccc@{}}
\toprule
\textbf{Method} & \textbf{Year} & \textbf{Camera} & \textbf{Lidar} & \textbf{4D Radar} & \textbf{Category} & \textbf{Assumptions} & \textbf{1 Frame} & \textbf{Open Dataset} & \textbf{Open-source} \\
\midrule
Scholler \textit{et al.} \cite{Scholler_radar_DL}       & 2019 & \checkmark & \xmark & \checkmark & end-to-end DL          & specific environment, training data, rotation only & \checkmark & \xmark & \xmark \\
Heng \cite{Heng_lidar_radar}          & 2020 & \xmark     & \checkmark & \checkmark & explicit geometry + optimization & sensor motion                                 & \xmark     & \xmark & \xmark \\
Persic \textit{et al.} \cite{Persic2020_radar}        & 2020 & \checkmark & \checkmark & 3D radar   & explicit geometry + optimization & dynamic objects, only yaw angle                      & \xmark     & nuScenes & \xmark \\
Cheng and Cao \cite{cheng2024onlinetargetlessradarcameraextrinsic}       & 2023 & \checkmark & \xmark     & \checkmark & DL + optimization & objects (pedestrian, car), training data                      & \xmark     & \xmark & \xmark \\
Wise \textit{et al.} \cite{wise_3d_radar_targetless}       & 2023 & \checkmark & \xmark     & \checkmark & explicit geometry + optimization & sensor motion                      & \xmark     & \xmark & \xmark \\
SensorX2Car \cite{sensorX2car}   & 2023 & \checkmark & \checkmark & 3D radar   & explicit geometry + optimization & sensor motion, only yaw angle                                 & \xmark     & \xmark & \checkmark \\

Ge \textit{et al.} \cite{segmentation_based_Radar_Ge}       & 2023 & \checkmark & \xmark & \checkmark & explicit geometry + optimization & stationary sensors, dynamic objects                                 & \xmark     & \xmark & \xmark \\

Hayoun \textit{et al.} \cite{Hayoun_clr_selfsup}        & 2024 & \checkmark & \checkmark & \checkmark & end-to-end DL          & dynamic objects, training data                & \checkmark & \xmark & \xmark \\
iKalibr \cite{iKalibr}       & 2025 & \checkmark & \checkmark & \checkmark & explicit geometry + optimization & sensor motion, IMU required                                & \xmark     & \xmark & \checkmark \\

Zhang  \textit{et al.} \cite{zhang_dl+pnp_Radar}      & 2025 & \checkmark & \xmark & 3D radar & DL + optimization & training data                                 & \xmark     & nuScenes & \xmark \\

4DRC-OC \cite{4DRC-OC}       & 2025 & \checkmark & \xmark     & \checkmark & end-to-end DL          & training data                                 & \checkmark & Dual-Radar & \checkmark \\
\textbf{CLRNet (ours)}  &      & \checkmark & \checkmark & \checkmark & end-to-end DL          & training data                                 & \checkmark & VoD and Dual-Radar  & \checkmark \\
\bottomrule
\end{tabular}
}
\end{table*}

\subsubsection{Targetless Calibration Involving 3D Radar}

Persic \textit{et al.} \cite{Persic2020_radar} use moving object tracking to perform 3D radar calibration; however, only the yaw angle is estimated, and prior knowledge of translation parameters is required.
SensorX2car \cite{sensorX2car} is a comprehensive open-source toolbox for sensor-to-vehicle frame calibration, supporting cameras, lidar, GNSS/INS, and 3D radar. However, they also ignore the translation and calibrate only the yaw angle of the rotational parameters of the radar sensor. Zhang 
\textit{et al.} \cite{zhang_dl+pnp_Radar} employs a three-branch neural network to extract features from images, 2D point clouds from radar, and radar cross-section (RCS) values, enabling robust cross-modal feature matching. A differentiable probabilistic PnP solver is then used for end-to-end pose estimation.

\subsubsection{Targetless Calibration Involving 4D Radar}

Scholler \textit{et al.} \cite{Scholler_radar_DL} proposes an end-to-end DL method for automatic targetless radar-camera calibration, but it only estimates the rotational component of the transform. They use Mobilenet 
to extract features from the image and max-pooling in the radar stream. However, the simplicity of the radar feature extraction layer and the lack of a dedicated matching layer limit the calibration accuracy.
Heng \cite{Heng_lidar_radar} calibrates radar to lidar sensors. The method first estimates lidar poses relative to the vehicle by minimizing point-to-plane distances between scans, then builds a 3D map used to determine the radar's 6-DoF pose via point-to-plane minimization, without requiring overlapping sensor fields of view.
Cheng and Cao \cite{cheng2024onlinetargetlessradarcameraextrinsic} use a DL based solution to extract common radar and camera features, and a PnP solver to solve their constructed optimization problem. 
Ge \textit{et al.} \cite{segmentation_based_Radar_Ge} performs initial calibration by constraining radar points within the contour of a segmented dynamic object. These are then refined by minimizing the velocity discrepancy between radar Doppler measurements and optical flow-derived camera velocities. The method's assumptions inherently limit the method's applicability to scenarios with stationary sensors and the presence of sufficiently dynamic objects, thereby narrowing its implementation scope.
Hayoun \textit{et al.} \cite{Hayoun_clr_selfsup}  propose a joint camera-lidar-radar targetless calibration approach, both with an optimization-based and a self-supervised DL based calibration. Their loss function is based on the distance of segmented object centers (after projection), the overlap of drivable areas, and a global consistency constraint. 

iKalibr \cite{iKalibr} is a continuous-time spatiotemporal calibration framework for IMUs, cameras, lidars, and radars. iKalibr performs calibration through a two-stage process: it first conducts a dynamic initialization to estimate initial spatiotemporal parameters and continuous-time motion trajectories, then refines these estimates using batch optimization based on a continuous-time B-spline representation of sensor motion. However, the method relies on 
sufficient large rotational and translational motion, which is often lacking in real-world vehicle operation, limiting its practical applicability.
Zhuang et al. propose 4DRC-OC \cite{4DRC-OC}, a calibration method for 4D radar and camera systems. Their approach introduces an auxiliary monocular depth estimation branch to align modality features and leverages dynamic convolution for adaptive feature extraction. A correlation module based on channel-wise fusion (CMCF) is used to enhance depth map matching. They perform experiments on the Dual-Radar dataset \cite{Dual-Radar}. In terms of our scope, 4DRC-OC is the closest related approach.

\section{Methodology}

\subsection{Problem Formulation and Data Preprocessing}

The goal of the calibration task is to provide the extrinsic parameters between the camera, lidar, and radar sensors using their recorded image and point cloud data as input.
This 6 DoF transformation can be represented by a $4\times4$ matrix $T=\mathcal{T}(q,t)$, that implements a rotation with quaternion $q$, and a 3D translation with vector $t$ in homogeneous coordinates.
Open datasets \cite{apalffy2022, Geiger2012CVPR_KITTI, Dual-Radar} usually provide multi-modal sensor data with fixed, and preliminary known extrinsic parameters \( T_{\text{fixed}} \), i.e., the transformation between the sensors remains the same during the whole dataset.
Therefore, to create a diversified dataset from \cite{apalffy2022} with large number of measurement frames having different extrinsic parameters, we perturb the initially constant \( T_{\text{fixed}} \) transformation with different \( T_{\text{mis}}(k) \) calibration parameters for each time frame $k$.
We simulate this miscalibration by transforming the original lidar and radar point cloud data with random translation and rotation parameters between predefined bounds, generating random values for Roll, Pitch, Yaw, X, Y, and Z. Our network is trained to regress \( T_{\text{mis}}(k) \), given time synchronized sensor data. 
The actual $T$ extrinsic parameters of a given time frame can be calculated as: \( T (k) = T_{\text{mis}}(k)\cdot T_{\text{fixed}}. \) For simpler notation, we will neglect the $k$ time index 
in the following.
    
\subsection{Equirectangular Projection}
\label{seq:projection}

To leverage the domain similarity between different sensors, we transform lidar and radar point cloud inputs into the depth image domain. Unlike existing DL based calibration methods \cite{LCCNet, CalibDepth, RGGNet, Scholler_radar_DL}, which rely on the pinhole camera model for projection, we propose using equirectangular projection.

When using the pinhole camera model, only a fixed frustum of the 3D space—corresponding to the camera's field of view (FoV)—is captured in the depth image. Due to the miscalibration described in the previous section (also similarly used by methods \cite{LCCNet, CalibDepth, RGGNet, Scholler_radar_DL}), some points that originally lay within the common FoV of the sensors are projected outside this frustum and thus discarded, resulting in information loss.
This approach yields richer and more informative depth images, which significantly improve the calibration process. The advantages of equirectangular projection are particularly evident in scenarios involving large-angle miscalibrations or when matching features between lidar and radar sensors. Since these sensors often have broader or differing FoV compared to cameras, retaining all recorded spatial measurements is crucial for achieving robust and accurate calibration. In addition, even under small miscalibrations or in pairwise setups, preserving the full radar scene leads to denser, spatially coherent depth maps, which allow the modality-specific encoder to extract more meaningful and robust features, ultimately improving calibration performance through better feature alignment and generalization. This is especially important for radar, given its sparsity and lower semantic richness compared to image or lidar data. In contrast, the equirectangular projection captures the full 360° FoV, ensuring that all points in the 3D space are preserved.

Given a point \((x, y, z)\) in Cartesian coordinates, we convert it to spherical coordinates (azimuth \(\theta\), elevation \(\phi\), radius \(r\)) and use equirectangular projection to map the normalized spherical coordinates ($\theta_{\text{norm}} = \frac{\theta + \pi}{2\pi}, \phi_{\text{norm}} = \frac{\phi + \frac{\pi}{2}}{\pi}$) to pixel coordinates \((u, v)\) of a given image with fixed width \(W\) and height \(H\): \( u = \left\lfloor \theta_{\text{norm}} \cdot W \right\rfloor, v = \left\lfloor (1 - \phi_{\text{norm}}) \cdot H \right\rfloor. \) On the first channel of the final lidar and radar depth images we have the point range ($r = \sqrt{x^2 + y^2 + z^2}$) information. On further channels, we store the additional inputs: reflection intensity for lidar and RCS, compensated velocity, and time for radar.

\subsection{Network Architecture}

\begin{figure*}[!t]
\centering
\includegraphics[width=\linewidth]{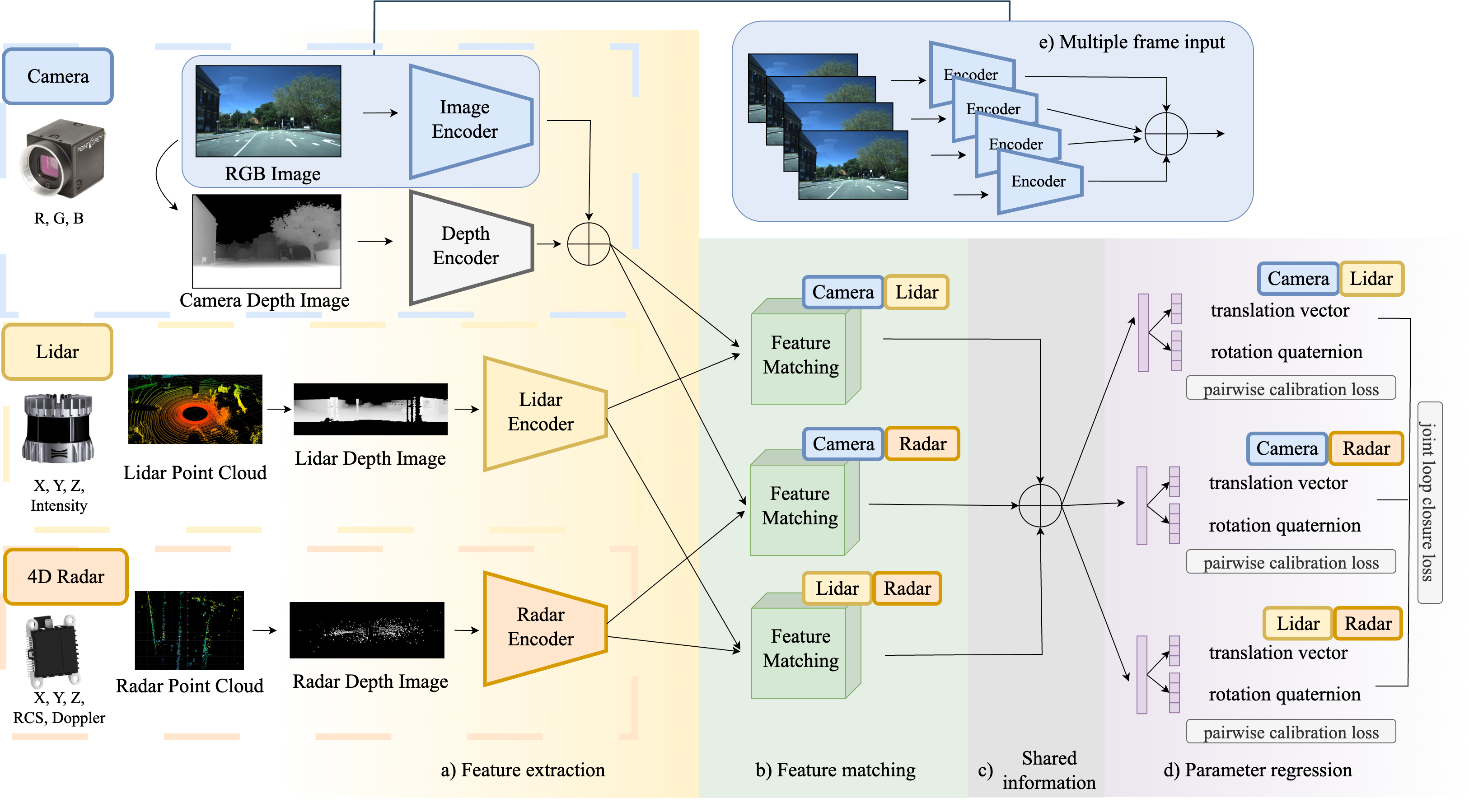}
\caption{CLRNet network architecture overview. On the lidar and radar branch, the input point clouds are projected using equirectangular projection before (a) feature extraction. After the pairwise (b) feature matching layers, we (c) concatenate the output to have shared information for the (d) parameter regression layers, which are used to predict the final extrinsic parameters for each three sensor pairs.}
\label{fig:CLRNet}
\end{figure*}

We propose CLRNet, a joint DNN architecture shown in Fig. \ref{fig:CLRNet}, for the multiple sensors to achieve robust and accurate calibration result by leveraging the shared information available from all sensors\footnote{The designed architecture including the scripts for training and testing, will be made freely available at https://github.com/tudelft-iv to academic and non-profit organizations for non-commercial, scientific use.}. First, we have individual branches to calculate feature maps from the sensor inputs, and then feature matching layers calculate the pairwise matching cost between all sensor pairs. Consequently, the resulting over-defined system can help to regress the final extrinsic parameters between the camera and radar more accurately than it would be possible with the use of the pairwise calibration architecture. Our architecture is inspired by LCCNet \cite{LCCNet}, a camera-lidar calibration DNN.

\subsubsection{Feature Extraction}

We utilize modality-specific encoders to extract high-level feature representations, enabling effective cross-modal alignment (see Fig. \ref{fig:CLRNet}(a)).
The camera image is passed through a ResNet-18 encoder. To enhance the network’s performance and guide it, as outlined in \cite{CalibDepth}, a depth map is predicted for each input image using DepthAnythingV2 \cite{depth_anything_v2}. This predicted depth map is then concatenated along the channel dimension with the feature map, after being processed by a modified ResNet-18 architecture incorporating Leaky ReLU activations.

The lidar and radar depth images (see Section \ref{seq:projection}) are going through a slightly modified Resnet-18 with LeakyReLU activation and multiple input channels: 2 channels for lidar (depth and intensity), 4 channels for radar (depth, RCS, compensated velocity [Doppler], and time).

To preserve fine spatial details essential for accurate calibration, we utilize a relatively large input size (512$\times$1024) for the encoders.

\subsubsection{Feature Matching}

The proposed architecture has three identical feature matching layers, one for each modality pair (see Fig. \ref{fig:CLRNet}(b)). The correlation between the two input feature maps is measured by a PWC-Net \cite{PWC-Net} inspired correlation layer. The layer is constructed as it was described in LCCNet \cite{LCCNet}, but using a larger input size (32$\times$64 instead of 12$\times$32) for improved feature matching.

\subsubsection{Parameter Regression from a Shared Feature Space}

The last layer is a custom, fully connected layer proposed for the required 3D transformation tasks. It processes an input tensor through several linear transformations, dropout regularization, and LeakyReLU activations, and outputs a translation vector and a rotation quaternion. The input is first projected to 512 neurons. Then the translation branch transforms this code to 256 neurons, and finally to 3 neurons representing the 3D translation vector, while the rotation branch transforms the input to 256 then to 4 neurons, representing the rotation quaternion.

We use three identical copies of the previously described layer to regress the transformation parameters between every sensor pair (see Fig. \ref{fig:CLRNet}(d)).
Instead of using as input to these regression layers separately the output from each previous feature matching layer, we propose using the concatenated feature vectors providing shared information from all sensors (Fig. \ref{fig:CLRNet}(c)). This step helps to enhance the accuracy of camera-radar calibration through the information from lidar.

\subsection{Loss Function}

Our proposed loss function \( \mathcal{L} \) combines two terms, the \textit{pairwise loss} and the \textit{joint loop closure loss}, balanced by  \( \lambda \):
\begin{equation}
    \mathcal{L} = (1-\lambda)\mathcal{L}^{\text{pairwise}} + \lambda \mathcal{L}^{\text{loop}}.
\end{equation}

\subsubsection{Pairwise Calibration Loss}

Between two sensors denoted by $i$ and $j$ the used calibration loss (\( \mathcal{L}_{ij}^{\text{pairwise}} \)) follows the approach of LCCNet \cite{LCCNet}: it is defined as the weighted (\( \lambda^{\text{pairwise}} \)) sum of two loss terms: parameter distance loss (\(\mathcal{L}_{ij}^{\text{param}} \)) and point cloud distance loss (\(\mathcal{L}_{ij}^{\text{point}} \)):

\begin{equation}
\mathcal{L}_{ij}^{\text{pairwise}} = (1 - \lambda^{\text{pairwise}}) \mathcal{L}_{ij}^{\text{param}} + \lambda^{\text{pairwise}} \mathcal{L}_{ij}^{\text{point}}.
\end{equation}

The parameter distance loss is defined as the difference between the predicted extrinsic parameters \(T_{ij}^{\text{pred}} =\mathcal{T} ( q_{ij}^{\text{pred}}, t_{ij}^{\text{pred}} \)) and the ground truth \(T_{ij}^{\text{GT}} =\mathcal{T} ( q_{ij}^{\text{GT}}, t_{ij}^{\text{GT}} \)). Smooth L1 loss \cite{LCCNet} is used due to its robustness to outliers, by calculating separately the angular distance \( L_q \) between rotation quaternions and the distance \( L_t \) between translation vectors:

\begin{equation}
\label{eq:param_loss}
\mathcal{L}_{ij}^{\text{param}} = \lambda^{\text{R}}L_q (q_{ij}^{\text{pred}}, q_{ij}^{\text{GT}}) + \lambda^{\text{T}} L_t (t_{ij}^{\text{pred}}, t_{ij}^{\text{GT}}),
\end{equation}
where \(\lambda^{\text{R}}\) and \(\lambda^{\text{T}}\) are scalar weights, as described in \cite{LCCNet}, used to address the difference in scale and convergence behavior between the two components.

The point cloud distance loss is determined by the discrepancy between the input point cloud transformed with the predicted and ground truth extrinsic parameters, respectively.
Given a point cloud \( P = \{p_1, p_2, \ldots, p_N\} \), with \( N \) points \( p_n \in \mathbb{R}^3 \), the point cloud distance loss \( \mathcal{L}_{ij}^{\text{point}} \) is defined as the mean average of Euclidean (L2) distances between points transformed with the predicted extrinsic matrix \(T_{ij}^{\text{pred}}\) and the ground truth extrinsic matrix \(T_{ij}^{\text{GT}}\):

\begin{equation}
\label{eq:point_loss}
\mathcal{L}_{ij}^{\text{point}}= \frac{1}{N} \sum_{n=1}^{N} \left\| T_{ij}^{\text{pred}} \cdot p_n - T_{ij}^{\text{GT}} \cdot p_n \right\|_2.
\end{equation}

The final pairwise loss is the sum of the individual pairwise losses between each sensor pair $i, j$:

\begin{equation}
\label{eq:pairwise}
\mathcal{L}_{\text{pairwise}} = \sum_{\substack{i,j \\ i < j}} (\mathcal{L}_{ij}^{\text{point}} + \mathcal{L}_{ij}^{\text{param}}),
\end{equation}
where $i$ and $j$ as indices represent two of the three available sensors (camera, lidar, radar).

\subsubsection{Joint Loop Closure Loss}
\label{seq:loop}

Besides the previously described pairwise calibration losses between the sensor pairs, we introduce a loss called the joint loop closure loss to encourage the network to have consistency among the pairwise calibration predictions. Similar approaches were used in other target-based and self-supervised calibration methods \cite{domhof_calibration, Hayoun_clr_selfsup} as well.
The individual sensors can be represented as vertices, and the pairwise transformation predictions between them can be represented as edges in a fully connected graph. The introduced loss measures the inconsistency between the edges in a loop, namely, applying consecutive transformations inside a loop would result in an identity matrix \( T^I \) (no transformation). As soon as we are using 3 sensors, we have one loop \( T^\text{loop} \) in our graph, which is the transformation from the camera (C) to the lidar (L) to the radar (R) and back to the camera: \( T^\text{loop} = T^\text{CL} \cdot T^\text{LR} \cdot T^\text{RC} \).

Afterward, the loop closure loss \( \mathcal{L}_{\text{loop}} \) can be calculated straightforwardly as the pairwise loss (Eq. \eqref{eq:param_loss}--\eqref{eq:pairwise}), but now using the distance between \( T^\text{loop} = \mathcal{T} (q^\text{loop}, t^\text{loop}) \) and \( T^I = \mathcal{T} (q^I, t^I) \), instead of \( T^{\text{pred}} \) and \( T^{\text{GT}} \). We show in Sec. \ref{seq:abl} that using the loop closure loss results in more accurate calibration and increased robustness to outliers, compared to relying solely on pairwise losses. Additionally, the presented loop closure loss scheme can be easily extended to handle more than three sensors.

\subsection{Simplified Network for Pairwise Calibration}

While the introduced CLRNet model expects synchronous camera, lidar ,and radar sensor inputs, its network architecture is largely modular (see Fig. \ref{fig:CLRNet}), which enables the construction of network variants for sensing platforms where only two (or more than three) sensors are available in a straightforward way. Apart from the analysis of the CLRNet, we will also evaluate a smaller network variant, called CRNet, for pairwise camera-radar calibration.
CRNet is assembled from a subset of the originally proposed CLRNet layers, including the corresponding feature encoders, the camera-radar feature matching module, and a parameter regression layer. The loss function of CRNet solely uses the pairwise loss term. 
Through the evaluation of  CRNet, we will show that our proposed three-sensor calibration approach exhibits a graceful degradation when the lidar sensor becomes inaccessible, and the advantages of various model features will be demonstrated in different configurations.





\subsection{Multi-frame  Calibration  for Rigid Sensor Systems}
\label{seq:mi}

Our previously described calibration workflow can separately estimate the extrinsic parameters for each time frame of the measurement sequence. This approach is necessary in cases of a non-rigid multi-sensor platform, where the calibration parameters might change in time (or even frame-by-frame). However, in several real-world applications, fully rigid sensor systems are utilized. As a result, we can assume that each frame is associated with the same extrinsic parameters; thus, we may use multiple frames as input to the network to regress a static extrinsic matrix. In our model version designed for rigid sensor platform scenarios, each branch receives multiple consecutive frames as an input, each input frame's features are extracted separately and the produced feature maps within the same branch are concatenated afterwards to produce a more descriptive shared representation of the captured features over multiple frames (see Fig. \ref{fig:CLRNet}(e)).
Furthermore, assuming a rigid platform allows us to propose a calibration pipeline that produces the final calibration result as the median (or mean) value from several predicted parameters over a complete measurement sequence, which step can further decrease the final calibration error, as demonstrated later in Section \ref{seq:results}. In our experiments, a 4-frame model variant proved to be the best, which we will henceforth refer to as CLRNet+4.

\section{Experiments}

\subsection{Datasets}

Several DL based camera-lidar calibration methods test and compare their results using the KITTI Odometry dataset \cite{Geiger2012CVPR_KITTI}, which, however, does not contain radar measurements; therefore, we use the View-of-Delft \cite{apalffy2022}, and Dual-Radar \cite{Dual-Radar} datasets for our experiments.

\subsubsection{View-of-Delft dataset}

For most of our experiments, we use the View-of-Delft dataset \cite{apalffy2022} with synchronized and calibrated lidar, camera, and 4D radar data acquired in complex, urban traffic.
The dataset uses a ZF FRGen21 4D radar providing approximately 400 points per frame. For the calibration problem, we only need the sensor data and the ground truth calibration parameters; therefore, we can use the available extended dataset with 29635 frames, so that the first 25000 frames are used for training, then the next 3300 frames for validation, and the last 1335 frames for testing. In the dataset, the ground truth parameters were obtained using \cite{domhof_calibration}, a target-based calibration method, performed with high attention to precise point selection.

\subsubsection{Dual-Radar dataset}

For comparison with 4DRC-OC \cite{4DRC-OC}, we use the recent Dual-Radar dataset \cite{Dual-Radar}, containing 10007 frames. Although the dataset setup includes a lidar sensor, the corresponding measurements were not extractable in the data we received from the authors. Consequently, our experiments are solely based on the camera and radar data. Consistent with \cite{4DRC-OC}, we use the Arbe Phoenix 4D radar from the dataset, which delivers high-resolution point clouds ranging from 6,000 to 14,000 points per frame.

\subsection{Implementation Details}

As network input, we use the camera images, lidar point clouds, and the 5-frame accumulated and ego-motion compensated radar point clouds \cite{apalffy2022}.
The point clouds are projected using equirectangular projection to a 1024$\times$2048 depth image, which is resized using linear interpolation to a 512$\times$1024 tensor used as input for the encoders. We train and test our models using a single NVIDIA GeForce GTX 1080 Ti with a batch size of 16 and Adam Optimizer. After trying different configurations, we had the best results using an initial learning rate of \( 1e^{-4}\), \(\lambda = 0.25\), \(\lambda^{\text{R}} = 1.0\), \(\lambda^{\text{T}} = 2.0\) and \( \lambda^{\text{pairwise}} = 0.5\). All networks were trained until convergence (15-100 epochs).

\subsection{Evaluation Metrics}
\label{seq:metrics}

The experimental results are analyzed based on the error between the predicted parameters and the ground truth. We used quaternions to represent the rotations, since they are unique and unambiguous. Similar to prior work \cite{4DRC-OC, LCCNet, CalibDepth, RGGNet, Scholler_radar_DL}, the error is represented with the angular (quaternion) and translational distance between the predicted and true values, for each prediction. Using the whole test set (1335 frames), we calculate the mean average error (MAE) and the median error for the translational (transl.) and rotational (rot.) parameters, respectively \cite{LCCNet, calibration_review}. For the calibration pipeline assuming rigid sensor platforms, we simulate miscalibration in the test set with 50 different random extrinsic parameters, and calculate the previously described average error metrics over the 50 experiments.

\subsection{Ablation Studies on the VoD Dataset}
\label{seq:abl}

To separately investigate the effects of the introduced model features, we tested our network in multiple configurations and reported the observed errors in Table \ref{tab:ablation1}.

\begin{table}[h]
\centering
\caption{Mean average camera-radar calibration error using the VoD dataset, \( \pm 20 \, \text{cm}\) and \( \pm 1 \, \text{°} \) miscalibration, and different configurations.}
\begin{tabular}{@{}lcc@{}}
\hline
\multirow{2}{*}{Model configuration} & \multicolumn{2}{c}{camera - radar MAE} \\
& Transl. (cm) & Rot. (°) \\ \hline
CRNet vanilla& 20.0 & 0.8\\
CRNet vanilla + equirectangular projection  & 14.0 & 0.6 \\
CRNet vanilla + 5-frame-input & 18.2 & 0.7 \\
CRNet vanilla + additional radar channels  & 16.6 & 0.6 \\
CRNet vanilla + camera depth branch  & 18.8 & 0.9 \\
CRNet (used all features) & 11.3 & 0.6 \\
CLRNet + loop closure loss & 9.5 & 0.5  \\
CLRNet + shared feature space & \textbf{9.1} & \textbf{0.4} \\
Proposed CLRNet (all features)& \textbf{9.1} & \textbf{0.4} \\
\hline
\end{tabular}
\label{tab:ablation1}
\end{table}

To isolate the effects of specific configurations on accuracy, we use CRNet to perform experiments on model components unrelated to lidar, thereby avoiding confounding influences from the lidar modality.
The first row of Table \ref{tab:ablation1} corresponds to CRNet vanilla, which uses pinhole projection, and only camera and radar depth input (without any additional channels). The following rows of the table represent different CRNet/CLRNet variants, where the added features are indicated. We can observe a significant performance improvement between CRNet vanilla and the complete CLRNet model, supporting the effectiveness of our added features. In particular, projecting the input point clouds using the equirectangular projection not only helps the matching between the lidar and radar, but leads to significant performance gain even in the CRNet setting (camera–radar only), demonstrating how preserving the full radar FoV improves feature extraction and matching quality of the network. More accurate translational calibration results were obtained using the 5-frame accumulated and ego motion compensated radar point cloud as input than in the vanilla approach, as the temporal accumulation increases point cloud density and scene coverage, providing richer spatial structure for feature extraction and alignment.
Using additional radar channels helps the network capture important information from the scene and improve the accuracy of the calibration. While using the predicted depth map from the camera image input did not improve the calibration accuracy as reported in \cite{CalibDepth, 4DRC-OC}, the translational error has indeed decreased.


Furthermore, the experiments show that incorporating lidar and adopting the CLRNet architecture significantly reduces the mean calibration error between the camera and radar---from \(11.3\,\text{cm},\, 0.6^\circ\) to \(9.1\,\text{cm},\, 0.4^\circ\). This improvement is primarily attributed to the use of regression layers operating on a shared (concatenated) feature space.

While applying loop closure loss alone improves accuracy compared to CRNet, combining it with the shared feature representation yields minimal additional benefit (only a few percent). This suggests that the shared feature space implicitly encodes information about parameter consistency across the sensor loop, thereby diminishing the standalone impact of the loop closure constraint.

\subsection{Results on the VoD dataset}
\label{seq:results}

\begin{table*}[h]
\centering
\setlength{\tabcolsep}{4pt}
\caption{Comparison of our proposed methods to SOTA reference methods on the VoD dataset. Mean average and median calibration error in different calibration scenarios. Values following the symbol \(\pm\) denote the 95\% confidence interval.}
\begin{tabular}{@{}ccccccccccc@{}}
\hline
\multirow{3}{*}{\shortstack{Calibration \\ Scenario}} & \multirow{3}{*}{Method} & \multicolumn{4}{c}{Camera - Radar Error} & & \multicolumn{4}{c}{Camera - Lidar Error}\\
\cline{3-6}
\cline{8-11}
& & \multicolumn{2}{c}{Translation (cm)} & \multicolumn{2}{c}{Rotation (°)} & & \multicolumn{2}{c}{Translation (cm)} & \multicolumn{2}{c}{Rotation (°)}\\ 
& & mean & median & mean & median & & mean & median & mean & median \\
\hline

\multirow{6}{*}{\shortstack{non-rigid system \\ one model \\\([\pm 20 \, \text{cm,} \pm 1 \, \text{°}]\)}} 
& Scholler et al. \cite{Scholler_radar_DL}\( ^1 \) & - & - & 1.0 $\pm$ 0.0 & 1.0 & & - & - & - & - \\
& Method based on \cite{Hayoun_clr_selfsup}\( ^1 \) & 18.9 $\pm$ 0.4 & 18.6 & 0.8 $\pm$ 0.0 & 0.7 & & 8.8 $\pm$ 0.3 & 7.6 & 0.5 $\pm$ 0.0 & 0.4\\
& 4DRC-OC \cite{4DRC-OC}\( ^2 \) & 18.5 $\pm$ 0.3 & 18.6 & 0.8 $\pm$ 0.0 & 0.8 & & - & - & - & - \\
& LCCNet \cite{LCCNet}\( ^2 \)& - & - & - & - & & 8.5 $\pm$ 0.2 & 6.8 & 0.3 $\pm$ 0.0 & 0.2 \\
& CRNet (ours) & 11.3 $\pm$ 0.3 & 10.0 & 0.6 $\pm$ 0.0 & 0.5 & & - & - & - & - \\
& CLRNet (\textbf{ours}) & \textbf{9.1} $\pm$ 0.3 & \textbf{7.8} & \textbf{0.4} $\pm$ 0.0 & \textbf{0.4} & & \textbf{4.7} $\pm$ 0.2 & \textbf{3.7} & \textbf{0.1} $\pm$ 0.0 & \textbf{0.1} \\
\hline

\multirow{6}{*}{\shortstack{non-rigid system \\ iterative refinement \\\([\pm 100 \, \text{cm,} \pm 20 \, \text{°}]\)}} 
& Scholler et al. \cite{Scholler_radar_DL}\( ^1 \) & - & - & 16.6 $\pm$ 0.3 & 16.5 & & - & - & - & - \\
& Method based on \cite{Hayoun_clr_selfsup}\( ^1 \) & 117 $\pm$ 3.2 & 107 & 9.7 $\pm$ 0.6 & 2.7 & & 50.8 $\pm$ 2.9 & 26.5 & 5.9 $\pm$ 0.4 & 1.6\\
& 4DRC-OC \cite{4DRC-OC}\( ^2 \) & 76.2 $\pm$ 1.6 & 75.2 & 3.9 $\pm$ 0.2 & 5.5 & & - & - & - & - \\
& LCCNet \cite{LCCNet}\( ^2 \)& - & - & - & - & & 17.4 $\pm$ 1.0 & 11.5 & 1.9 $\pm$ 0.2 & 0.3\\
& CRNet(ours) & 14.7 $\pm$ 0.6 & 12.4 & 0.8 $\pm$ 0.0 & 0.7 & & - & - & - & - \\
& CLRNet (\textbf{ours}) & \textbf{12.0} $\pm$ 0.5 & \textbf{10.0} & \textbf{0.5} $\pm$ 0.0 & \textbf{0.4} & & \textbf{4.8} $\pm$ 0.3 & \textbf{3.5} & \textbf{0.1} $\pm$ 0.0 & \textbf{0.1} \\
\hline

\multirow{7}{*}{\shortstack{rigid system \\ iterative refinement \\\([\pm 100 \, \text{cm,} \pm 20 \, \text{°}]\)}} 
& Scholler et al. \cite{Scholler_radar_DL}\( ^1 \) & - & - & 16.7 $\pm$ 2.1 & 16.7 & & - & - & - & - \\
& Method based on \cite{Hayoun_clr_selfsup}\( ^1 \) & 112 $\pm$ 19.7 & 105 & 8.6 $\pm$ 4.2 & 1.0 & & 61.4 $\pm$ 21.2 & 33.1 & 5.0 $\pm$ 2.3 & 0.5\\
& 4DRC-OC \cite{4DRC-OC}\( ^2 \) & 67.5 $\pm$ 10.5 & 72.5 & 2.2 $\pm$ 1.4 & 0.9 & & - & - & - & - \\
& LCCNet \cite{LCCNet}\( ^2 \)& - & - & - & - & & 7.2 $\pm$ 5.2 & 1.8 & 1.5 $\pm$ 1.3 & 0.1 \\
& CRNet (ours) & 1.9 $\pm$ 0.4 & 1.8 & 0.2 $\pm$ 0.02 & 0.2 & & - & - & - & - \\
& CLRNet (ours) & 1.8 $\pm$ 0.1 & 1.7 & 0.1 $\pm$ 0.03 & 0.1 & & \textbf{0.2} $\pm$ 0.0 & \textbf{0.2} & 0.1 $\pm$ 0.0 & 0.1 \\
& CLRNet+4 (\textbf{ours}) & \textbf{0.9} $\pm$ 0.0 & \textbf{0.9} & \textbf{0.1} $\pm$ 0.0 & \textbf{0.1} & & 0.3 $\pm$ 0.01 & 0.3 & \textbf{0.0} $\pm$ 0.0 & \textbf{0.03} \\
\hline

\multirow{7}{*}{\shortstack{rigid system \\ iterative refinement \\ \([\pm 200 \, \text{cm,} \pm 180 \, \text{°}]\)}}
& Scholler et al. \cite{Scholler_radar_DL}\( ^1 \) & - & - & 122 $\pm$ 10.9 & 113 & & - & - & - & - \\
& Method based on \cite{Hayoun_clr_selfsup}\( ^1 \) & 285 $\pm$ 41.3 & 277 & 124 $\pm$ 12.8 & 118 & & 280 $\pm$ 37.2 & 273 & 125.0 $\pm$ 16.7 & 125\\
& 4DRC-OC \cite{4DRC-OC}\( ^2 \) & 182 $\pm$ 21.9 & 168 & 132 $\pm$ 17.2 & 145.0 & & - & - & - & - \\
& LCCNet \cite{LCCNet}\( ^2 \)& - & - & - & - & & 186 $\pm$ 40.5 & 192 & 99.2 $\pm$ 28.8 & 130 \\
& CRNet (ours) & 13.4 $\pm$ 11.3 & 4.2 & 1.1 $\pm$ 1.1 & 0.3 & & -  & - & - & - \\
& CLRNet (ours) & 3.3 $\pm$ 1.0 & 2.1 & 0.3 $\pm$ 0.2 & \textbf{0.2} & & \textbf{0.4} $\pm$ 0.1 & 0.3 & 0.1 $\pm$ 0.01 & 0.1 \\
& CLRNet+4 (\textbf{ours}) & \textbf{3.2} $\pm$ 1.5 & \textbf{1.4} & \textbf{0.3} $\pm$ 0.1 & 0.2 & & 0.4 $\pm$ 0.2 & \textbf{0.2} & \textbf{0.1} $\pm$ 0.04 & \textbf{0.04} \\
\hline

\multicolumn{11}{l}{\raggedright \( ^1 \) Re-implementations of the original models retrained on VoD, see Section \ref{seq:results}.}\\
\multicolumn{11}{l}{\raggedright \( ^2 \) Original models retrained on VoD, see Section \ref{seq:results}.}
\end{tabular}
\label{tab:results}
\end{table*}

Table \ref{tab:results} summarizes results across four test scenarios: (1) single-frame model for small miscalibrations, (2) iterative refinement for large miscalibrations, (3) rigid setup with temporal model, and (4) full-range miscalibration recovery. Below, we briefly highlight performance in each case.

First, we evaluated the output of a single DNN model that was trained and tested with single-frame measurement inputs with a miscalibration range of  \( \pm 20  \, \text{cm} \) and \( \pm 1\, \text{°}\). Here, our CLRNet model yielded a median camera - radar error of \( 7.8 \, \text{cm}, 0.4  \, \text{°} \) and a median camera - lidar error of \( 3.7 \, \text{cm}, 0.1  \, \text{°} \), as shown in the upper part of Table \ref{tab:results}.

Second, for processing measurement frames with large miscalibrations \( (\pm 100 \, \text{cm}, \pm 20  \, \text{°}) \), we adopt the sequential iterative refinement approach \cite{regnet, LCCNet}, consecutively running 4 DNN models trained with miscalibrations of \( [\pm 100 \, \text{cm}, \pm 20  \, \text{°}] \), \( [\pm 50 \, \text{cm}, \pm 5  \, \text{°}] \), \( [\pm 20 \, \text{cm}, \pm 1  \, \text{°}] \), \( [\pm 5 \, \text{cm}, \pm 0.5  \, \text{°}] \), respectively. Here, each subsequent model's input is the calibration prediction of the previous model. 
In this scenario we can achieve a median camera - radar error of \( 10.0 \, \text{cm}, 0.4  \, \text{°} \) and a median camera - lidar error of \( 3.5 \, \text{cm}, 0.1  \, \text{°} \), which demonstrates that with applying the iterative refinement framework the method can adjust large miscalibrations without a significant loss of accuracy compared to the first test scenario's simpler configuration.


Third, in the scenario assuming a rigid sensor setup, we can greatly increase the calibration accuracy using a calibration pipeline, resulting in the median extrinsic parameter over the whole test set (see Section \ref{seq:mi} and \ref{seq:metrics}). In this scenario, we can also use our more advanced network, utilizing temporal information using multiple frames as input. During our experiments with different numbers of frames, we had the best results using four consecutive frames as input for the multiple input network, called CLRNet+4. Using CLRNet+4 as the last model during iterative refinement and the above-mentioned calibration pipeline, we yielded a median camera - radar error of \( 0.9 \, \text{cm}, 0.1  \, \text{°} \) and a median camera - lidar error of \( 0.3 \, \text{cm}, 0.03  \, \text{°} \).

Lastly, in the fourth scenario we test our method similarly as in the third scenario (rigid system, iterative refinement) but a \( [\pm 200 \, \text{cm}, \pm 180  \, \text{°}] \) miscalibration was applied to the input ensuring full coverage of the observable field. We trained an additional model using this \( [\pm 200 \, \text{cm}, \pm 180  \, \text{°}] \) miscalibration, and added it as the first model to the iterative refinement approach. Here, our CLRNet+4 method is capable of achieving a median camera - radar error of \( 1.4 \, \text{cm}, 0.2  \, \text{°} \) and a median camera - lidar error of \( 0.2 \, \text{cm}, 0.04  \, \text{°} \) (see the bottom part of Table \ref{tab:results}).
Although large synthetic miscalibrations differ from native sensor setups, they effectively demonstrate our model's robustness.

We perform comparison with the method \cite{Scholler_radar_DL}, \cite{Hayoun_clr_selfsup}, and \cite{4DRC-OC}, which are, in terms of goals and assumptions, the closest to our approach (end-to-end DL based calibration solutions for sensor setups containing 4D radar). 
As code was unavailable for \cite{Scholler_radar_DL} and \cite{Hayoun_clr_selfsup}, we reimplemented both based on paper descriptions. For \cite{Hayoun_clr_selfsup}, lacking architecture details, we used a supervised DNN with a ResNet backbone and standard FC layers to regress extrinsics.

In the four experimental scenarios introduced earlier, we found that the baseline radar calibration techniques—when retrained on the VoD dataset—consistently yielded substantially lower calibration accuracy than our proposed method, with errors often one to two orders of magnitude higher. This significant performance gap is most likely due to two key factors, which are elaborated in the Discussion section (Section V). The results are shown in Table \ref{tab:results}. For example, in the first scenario 
our method was able to achieve a median \( 7.8\, \text{cm}, 0.4\, \text{°} \) calibration error between the camera and radar, and \( 3.7 \, \text{cm}, 0.1\, \text{°} \) for camera-lidar, while the method \cite{Scholler_radar_DL} produced a rotational error of around $1.0\text{°}$ in the VoD dataset,
the method based on \cite{Hayoun_clr_selfsup} achieved a median \( 18.6 \, \text{cm}, 0.7\, \text{°} \) camera-radar, and \( 7.6 \, \text{cm}, 0.4\, \text{°} \) camera-lidar calibration error, and the method \cite{4DRC-OC} achieved a median \( 18.6 \, \text{cm}, 0.8\, \text{°} \) camera-radar error.
Our experiments showed similar tendencies in calibration accuracy in the other scenarios as well (see Table \ref{tab:results}).

We also compare our approach to our camera-lidar calibration baseline LCCNet \cite{LCCNet}. The LCCNet model retrained on VoD using \( \pm 20  \, \text{cm} \) and \( \pm 1\, \text{°}\) miscalibration could achieve a median \( 6.8\, \text{cm}, 0.2\, \text{°} \) calibration error between the camera and lidar having around a 40 \% worse accuracy than CLRNet (ours). We attribute this improvement to the fact that we used equirectangular projection, an additional intensity channel from lidar, a predicted depth image in the camera branch, and bigger input tensors (512$\times$1024 instead of 256$\times$512) to capture fine spatial details.

Additionally, we performed a calibration experiment using the iKalibr framework \cite{iKalibr}, a traditional motion-based, targetless approach. Although the VoD dataset comprises diverse scenes typical of autonomous vehicle operation, the calibration process failed, reporting insufficient excitation in sensor motion. As also noted in the original paper introducing iKalibr, successful calibration relies on "sensors that undergo sufficiently excited motion", highlighting the importance of rich motion dynamics in enabling accurate parameter estimation. This result underscores a key limitation of traditional calibration methods: their strong dependence on specific data characteristics—such as adequately excited motion trajectories—which, if absent, render the calibration either infeasible or unreliable.

\begin{figure}[t]
\centering
\includegraphics[width=\linewidth]{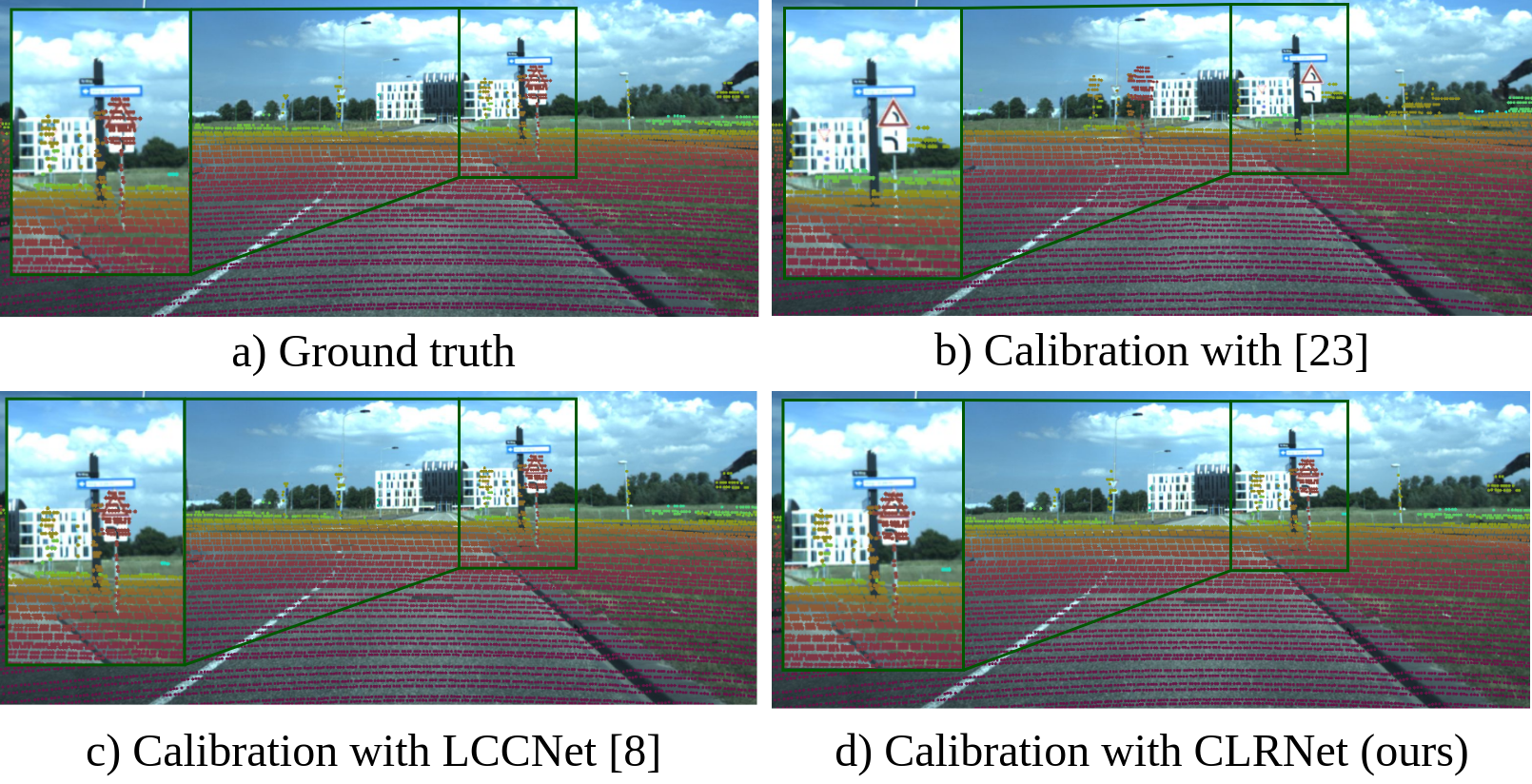}
\caption{Qualitative results of lidar point projection to the camera image with two reference methods \cite{LCCNet,Hayoun_clr_selfsup} and the proposed CLRNet model. CLRNet is superior to \cite{Hayoun_clr_selfsup}, also using three sensors; while our approach similarly performs to \cite{LCCNet}, which is a pairwise lidar-camera calibration baseline method. }
\label{fig:qal_lidar}
\end{figure}

\begin{figure*}[!t]
\centering
\includegraphics[width=0.95\linewidth]{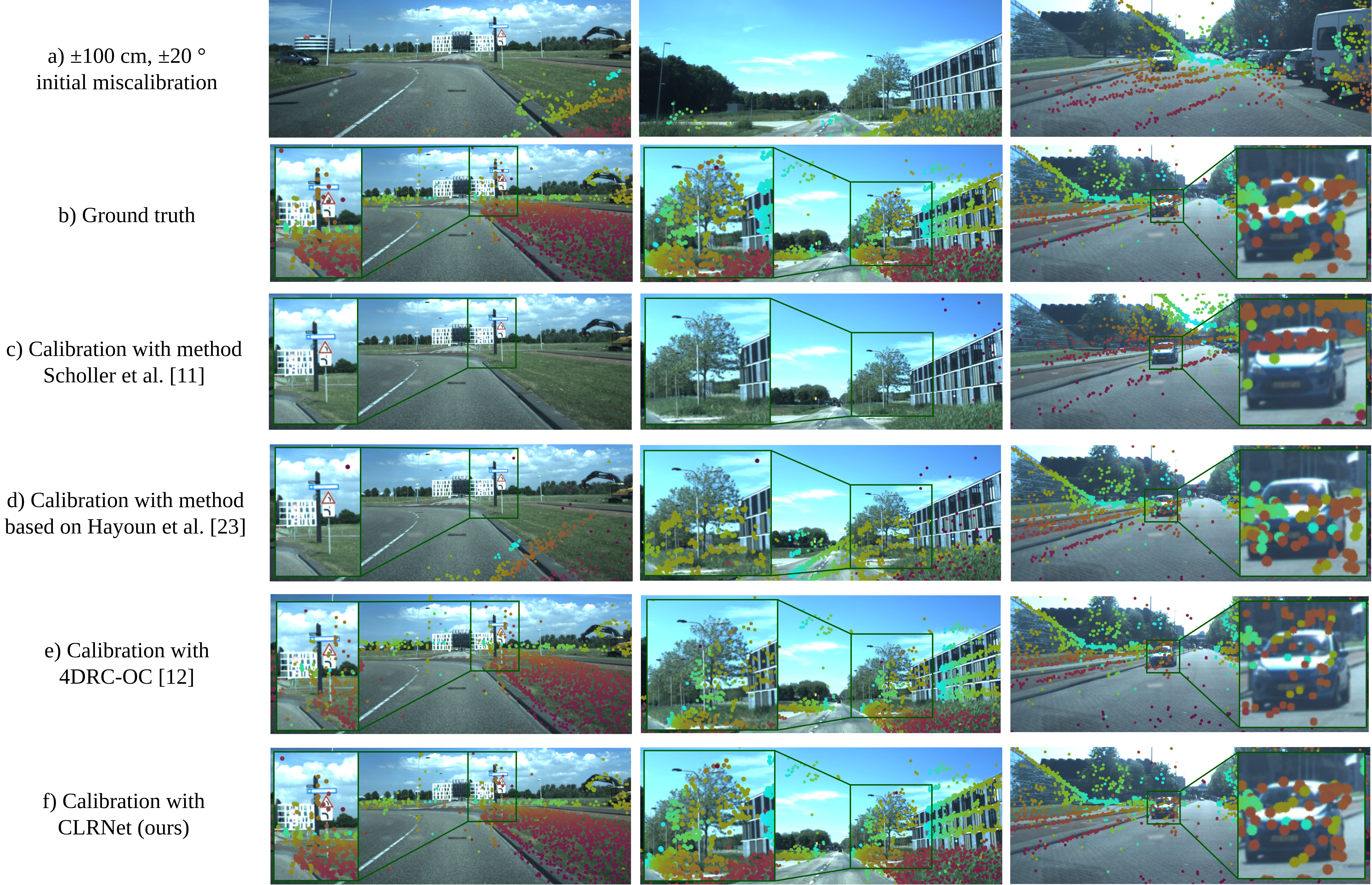}
\caption{Qualitative results of radar point projection to the camera image in three test scenes by using the calibration parameters of the (b) ground truth, (c)-(e) the three reference methods \cite{Scholler_radar_DL, Hayoun_clr_selfsup, 4DRC-OC} and (f) the proposed CLRNet model.}
\label{fig:qal_radar}
\end{figure*}

Figures \ref{fig:qal_lidar} and \ref{fig:qal_radar} show lidar and radar points projected onto camera images using extrinsic parameters obtained from different calibration methods in the second scenario - a non-rigid system with iterative refinement, correcting \( [\pm 100 \, \text{cm}, \pm 20 \, \text{°} ]\) (see Section \ref{seq:results}). The results indicate that our CLRNet predictions achieve more accurate point alignment with reference objects in the camera image compared to other methods.

\subsection{Results on Domain Transfer (VoD and Dual-Radar)}
\label{seq:dual_exp}

We now assess in-domain performance (training and testing on the same dataset) and cross-domain generalization (training on one dataset and testing on the other) for the two datasets Dual-Radar \cite{Dual-Radar} and VoD \cite{apalffy2022}. The results are summarized in Table~\ref{tab:generalization_matrix} for the baseline 4DRC-OC \cite{4DRC-OC} and our proposed CRNet.

In the in-domain case on Dual-Radar, CRNet achieves a median calibration error of (\( 1.8\,\text{cm}, 0.2^{\circ} \)), outperforming 4DRC-OC (\( 3.8\,\text{cm}, 0.2^{\circ} \)). Similarly, when trained and tested on VoD, CRNet shows clear improvements over the baseline.

In the cross-domain case, both methods show a remarkable performance drop. Still, CRNet manages to retain lower errors than 4DRC-OC. For example, when trained on VoD and tested on Dual-Radar, CRNet achieves (\( 139\,\text{cm}, 6.3^{\circ} \)) compared to 4DRC-OC’s (\( 186\,\text{cm}, 8.3^{\circ} \)). We revisit domain transfer capability in the next Section. 


\begin{table}[h]
\centering
    \caption{Domain transfer results for 4DRC-OC \cite{4DRC-OC} and CRNet (ours) involving the Dual-Radar and VoD datasets. Numbers show mean / median / std MAE for translation (cm) and rotation (°), correcting  $[\pm 1.0\,\text{m}, \pm 20^\circ]$ miscalibrations using iterative refinement and assuming rigid platforms.}
\setlength{\tabcolsep}{3pt}
\renewcommand{\arraystretch}{1.2}
\begin{tabular}{@{}c|c|c|c|c@{}}
\hline
\multirow{2}{*}{Method} & \multirow{2}{*}{Train} & \multirow{2}{*}{Test} & \multicolumn{2}{c}{camera–radar MAE} \\
 & & & Transl. (cm) & Rot. (°) \\ \hline

\multirow{4}{*}{4DRC-OC} 
& VoD  & Dual & 170 / 186 / 43 & 9.0 / 8.3 / 2.3 \\
& Dual & Dual & 4.8 / 3.8 / 4.5 & 0.4 / 0.2 / 0.4 \\
& Dual & VoD  & 190 / 194 / 55 & 7.0 / 6.6 / 1.5 \\ 
& VoD  & VoD  & 67.5 / 72.5 / 26.8 & 2.2 / 0.9 / 3.6 \\ \hline 

\multirow{4}{*}{CRNet (ours)} 
& VoD  & Dual & 134 / 139 / 46.8 & 6.5 / 6.3 / 1.3 \\
& Dual & Dual & 1.8 / 1.8 / 0.4 & 0.2 / 0.2 / 0.1 \\
& Dual & VoD  & 196 / 203 / 49.8 & 9.6 / 9.6 / 0.6 \\ 
& VoD  & VoD  & 1.9 / 1.8 / 1.0 & 0.2 / 0.2 / 0.1 \\ 




\hline
\end{tabular}
\label{tab:generalization_matrix}
\end{table}

\section{Discussion}
\label{seq:discussion}

The experiments show that the proposed network variants, CLRNet and CRNet, outperform baselines on both VoD \cite{apalffy2022} and Dual-Radar \cite{Dual-Radar}. They attain sub-degree rotation and centimeter-level translation errors—values commensurate with automated-driving fusion requirements reported in the literature (sub-degree/centimeter for camera–lidar and slightly looser for camera–radar) \cite{Tan2024,BEVCalib2025}. We attribute these gains to our architecture and design choices—equirectangular projection, additional input channels, an optional camera-depth branch, a shared feature space, and a joint loop-closure term—which significantly reduce calibration error in-domain (see Sec.~\ref{seq:projection}, Table~\ref{tab:ablation1}).

Cross-dataset generalization is challenging: when transferring between VoD and Dual-Radar, calibration error increases for all methods—including ours—reflecting a larger domain gap than typically studied before in the field of camera-lidar calibration \cite{han2025dfcalib,DXQ-Net,huang2025whatmatters,cmrnext}. Contributing factors include (i) a \mbox{15–30$\times$} difference in radar point-cloud density, (ii) non-identical camera placements and intrinsics, and (iii) distributional shifts in traffic and illumination (day-only vs. day- and night-time). Our method remains comparatively stronger under these shifts (Table~\ref{tab:generalization_matrix}). 
We expect robustness to increase as larger and more diverse tri-modal datasets become available.

Our approach is useful when one-shot (single-frame) calibration is required—for example, on non-rigid platforms subject to articulation or vibration (Sec.~\ref{seq:introduction}). It is also well-suited to fixed sensor suites (factory setups or research vehicles) where the same sensors are reused but occasionally remounted: accurate results can be obtained without retraining; retraining is only required when a new sensor type is introduced (requiring only ground-truth calibration parameters and some data, not per-sample manual annotation).

On an NVIDIA GeForce GTX 1080~Ti, a single CLRNet model—including the point-cloud–to–depth-image transformation but excluding camera image–based depth prediction—runs at 20 ms per frame (50 fps), sufficient for real-time use.

Further work may proceed in two directions: (i) improving domain transfer for methods involving 4D radar, and (ii) exploring unsupervised approaches for targetless calibration to further reduce annotation effort.

\section{Conclusion}


We proposed CLRNet, an end-to-end deep learning-based, targetless extrinsic calibration method for camera, lidar, and 4D radar. Through extensive ablation studies, we demonstrated how each component of our architecture—including the shared feature space, equirectangular projection, loop closure loss, camera-based depth input, and additional radar channels—contributes to improving calibration accuracy. Building on these insights, our full approach achieves state-of-the-art results on both the VoD and Dual-Radar datasets. In the rigid sensor setup correcting \([\pm 100 \, \text{cm,} \pm 20 \, \text{°}]\) miscalibrations, we achieved \(0.9 \, \text{cm}, 0.1\, \text{°}\) median error for camera–radar and \(0.3  \, \text{cm}, 0.03\, \text{°}\) for camera–lidar calibration, significantly outperforming existing baselines. Future work involves developing DL models, which can better transfer from one domain to another in the current supervised learning setting, and extending to the unsupervised learning case.


\bibliographystyle{IEEEtran}

\bibliography{bibliography}

\end{document}